\documentclass{bmvc2k}

\usepackage{epsfig}
\usepackage{graphicx}
\usepackage{amsmath}
\usepackage{amssymb}
\usepackage{colortbl}
\usepackage{stackengine}
\usepackage{multirow}
\usepackage{rotating}
\usepackage{xcolor, cancel}

\usepackage{float}
\usepackage{listings}
\newfloat{lstfloat}{htbp}{lop}
\floatname{lstfloat}{Algorithm}
\usepackage{pifont}
\usepackage{booktabs}
\usepackage{xcolor}

\definecolor{xarch}{rgb}{1.0, 0.9, 0.9}
\definecolor{sarch}{rgb}{0.94, 0.97, 1.0}
\definecolor{light_green}{rgb}{0.72, 0.85, 0.85}
\definecolor{light_red}{rgb}{1.0, 0.9, 0.9}

\DeclareRobustCommand{\rchi}{{\mathpalette\irchi\relax}}
\newcommand{\irchi}[2]{\raisebox{\depth}{$#1\chi$}}
\def\ie{\emph{i.e.}} 
\def\eg{\emph{e.g.}} 

\definecolor{codegreen}{rgb}{0,0.5,0}
\definecolor{codegray}{rgb}{0.5,0.5,0.5}
\definecolor{codepurple}{rgb}{0.58,0,0.82}
\definecolor{backcolour}{rgb}{1.0,1.0,1.0}
\definecolor{link}{rgb}{1.0,1.0,1.0}

\lstdefinestyle{mystyle}{
    backgroundcolor=\color{backcolour},   
    commentstyle=\color{codegreen},
    keywordstyle=\color{magenta},
    numberstyle=\tiny\color{codegray},
    stringstyle=\color{codepurple},
    basicstyle=\ttfamily\footnotesize,
    breakatwhitespace=false,         
    breaklines=true,                 
    keepspaces=true,                 
    numbers=left,                    
    numbersep=5pt,                  
    showspaces=false,                
    showstringspaces=false,
    showtabs=false,                  
    tabsize=1,
    float=tp,
  floatplacement=tbp
}
\definecolor{lightred}{rgb}{238,208,219}
\newcommand\redcancel[2][red]{\renewcommand\CancelColor{\color{#1}}\cancel{#2}}

\usepackage{mathtools}
\DeclarePairedDelimiterX{\infdivx}[2]{(}{)}{%
  #1\;\delimsize\|\;#2%
}

\DeclarePairedDelimiter{\norm}{\lVert}{\rVert}

\newcommand{\Real}{{\rm I\!R}}

\newcommand{\Renyi}{{R\'enyi}}

\newcommand{\functional}{function\;}  

\definecolor{comment}{rgb}{0.0627, 0.5647, 0.0235}

\newcommand*{\colorboxed}{}
\def\colorboxed#1#{%
  \colorboxedAux{#1}%
}
\newcommand*{\colorboxedAux}[3]{%
  \begingroup
    \colorlet{cb@saved}{.}%
    \color#1{#2}%
    \boxed{%
      \color{cb@saved}%
      #3%
    }%
  \endgroup
}

\newcounter{daggerfootnote}

\newcommand\blfootnote[1]{%
  \begingroup
  \renewcommand\thefootnote{}\footnote{#1}%
  \addtocounter{footnote}{-1}%
  \endgroup
}

\title{Information Theoretic Representation Distillation}

\addauthor{Roy Miles$^\ast$}{r.miles18@imperial.ac.uk}{1}
\addauthor{Adrian Lopez-Rodriguez$^\ast$}{al4415@imperial.ac.uk}{1}
\addauthor{Krystian Mikolajczyk}{k.mikolajczyk@imperial.ac.uk}{1}

\addinstitution{
MatchLab \\
Imperial College London \\
Department of Electrical and Electronic
Engineering \\
London, UK
}

\runninghead{Miles et al.}{Information Theoretic Representation Distillation}

\def\eg{\emph{e.g}\bmvaOneDot}

\begin{document}

\maketitle

\blfootnote{$^\ast$ The authors contributed equally to this paper}

\begin{abstract}
Despite the empirical success of knowledge distillation, current state-of-the-art methods are computationally expensive to train, which makes them difficult to adopt in practice. To address this problem, we introduce two distinct complementary losses inspired by a cheap entropy-like estimator. These losses aim to maximise the correlation and mutual information between the student and teacher representations. Our method incurs significantly less training overheads than other approaches and achieves competitive performance to the state-of-the-art on the knowledge distillation and cross-model transfer tasks. We further demonstrate the effectiveness of our method on a binary distillation task, whereby it leads to a new state-of-the-art for binary quantisation and approaches the performance of a full precision model. Code: \href{http://github.com/roymiles/ITRD}{\texttt{github.com/roymiles/ITRD}}

\end{abstract}

\section{Introduction}
Deep learning has significantly advanced state-of-the-art across a wide  range of computer vision tasks. Despite this success, most models are too computationally expensive to deploy on resource-constrained devices. Fortunately, the training of such models is coupled with significant parameter redundancy, which has been explicitly exploited in the pruning and quantisation literature~\cite{Zhuang2018Discrimination-awareNetworks, Li2017PruningConvnets, BethgeMeliusNet, Rastegari2016XNOR-Net:Networks}. Knowledge distillation proposes an alternative approach whereby a much larger pre-trained model can provide additional supervision for a smaller model during training. This paradigm removes the restriction of the two models to share the same underlying architecture, thus enabling hand-crafted designs of the target architecture to meet the imposed resource constraints. However, some of the recent state-of-the-art distillation methods, \eg~the recent union of self-supervision and knowledge distillation~\cite{Yang2021HierarchicalDistillation, Xu2020KnowledgeSelf-supervision},  have made it increasingly expensive to train these student models. To this end, we develop a distillation method with a low computational overhead.\par
Information theory provides a natural lens for quantifying the statistical relationship between these models, and so is a common framework for deriving distillation losses~\cite{Chen2020WassersteinDistillation, Tian2019ContrastiveDistillation}. 
Hence, we propose \textbf{I}nformation \textbf{T}heoretic \textbf{R}epresentation \textbf{D}istillation (ITRD) as a unified and computationally efficient framework that directly connects information theory with representation distillation. Specifically, this framework is inspired by the generalised \Renyi's entropy and makes the training for specific applications more effective. \Renyi's entropy is a generalisation of Shannon's entropy and has led to improvements in other areas~\cite{Mironov2017RenyiPrivacy,Yu2019UnderstandingConcepts,Shekar2011FaceAnalysis}.
\begin{figure}[t]
\centering
    \includegraphics[width=.95\linewidth]{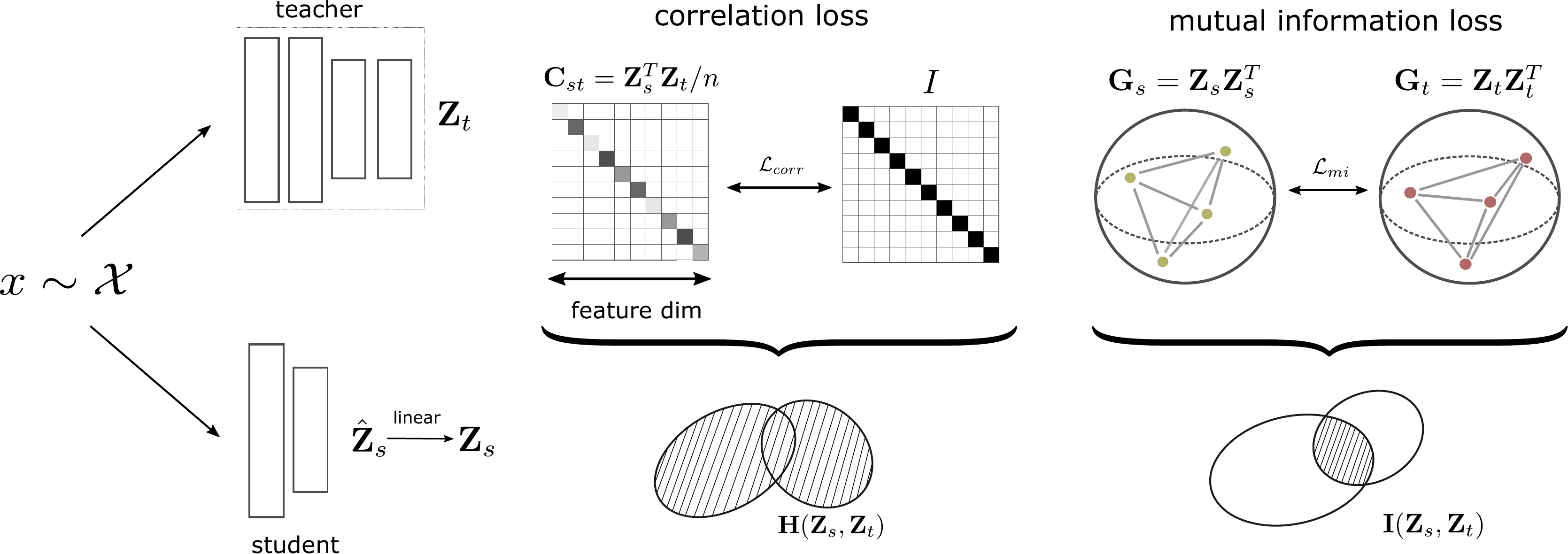}
    \vspace{1em}
    \caption{Information theoretic representation distillation (ITRD) involves two distinct losses, namely a correlation loss and a mutual information loss. The former loss maximises the correlation between the student and teacher, while the latter maximises a quantity resembling the mutual information that aims to transfer the intra-batch sample similarity.}
    \label{fig:overview}
\end{figure}
As figure \ref{fig:overview} shows, we propose to model the distillation task with two distinct loss functions that correspond to maximising the correlation and mutual information between the student and teacher representations. The correlation loss aims to increase the similarity between teacher and student representations across the feature dimension. Conversely, the mutual information loss aims to match the intra-batch sample similarity between the teacher and the student. Our results show a strong accuracy v.s. training cost trade-off 
in comparison to state-of-the-art across two standard benchmarks,  CIFAR100 and ImageNet, for a range of architecture pairings where we achieve up to $24.4\%$ relative improvement. Our loss directly addresses the training efficiency problem, which we believe will encourage its adoption amongst machine learning researchers and practitioners.
We further demonstrate the effectiveness of this framework on representation transfer, binary network transfer and NLP architecture transfer, whereby we are able to improve upon the state-of-the-art for all tasks. 

%
%
\vspace{-0.5em}
\section{Related Work}
\paragraph{Knowledge Distillation} (KD) attempts to transfer the knowledge from a large pre-trained model (teacher) to a much smaller compressed model (student). This was originally introduced in the context of image classification~\cite{Hinton2015DistillingNetwork}, whereby the soft predictions of the teacher can act as pseudo ground truth labels for the student. The soft predictions then provide the student with supervision on the correlations between classes which are not explicitly available from one-hot encoded ground truth labels. Spherical knowledge distillation~\cite{Guo2020ReducingDistillation} proposes to re-scale the logits before KD to address the capacity gap problem, while Prime-Aware Adaptive Distillation~\cite{ZhangPrime-AwareDistillation} introduces an adaptive sample weighting.
Hinted losses provide a natural extension of KD using an $L_2$ distance between the student and teacher's intermediate representation~\cite{Romero2015FitNets:Nets}. Attention transfer~\cite{Zagoruyko2019PayingTransfer} proposed to re-weight the spatial entries before the matching losses, while neuron selectivity transfer~\cite{Huang2017LikeTransfer}, similarity-preserving KD~\cite{Tung2019Similarity-preservingDistillation}, and relational KD~\cite{Park2019RelationalDistillation} attempt to transfer the structural similarity.
Similarly, FSP matrices~\cite{Yim2017ALearning} attempt to capture the flow of information and Review KD~\cite{ChenDistillingReview} propose the use of attention-based and hierarchical context modules. KD can also be modelled directly within a probabilistic framework~\cite{Ahn2019VariationalTransfer, Passalis2018LearningTransfer} through estimating and maximising the mutual information between the student and the teacher. ICKD~\cite{Liu2021ExploringDistillation} propose to transfer the correlation between channels of intermediate representations. 
A natural extension of supervised contrastive learning in the context of knowledge distillation was proposed in CRD~\cite{Tian2019ContrastiveDistillation}.  WCoRD~\cite{Chen2020WassersteinDistillation} also use a contrastive learning objective but through leveraging the dual and primal forms of the Wasserstein distance. CRCD~\cite{ZhuComplementaryDistillation} further develop this contrastive framework through the use of both feature and gradient information. Unfortunately, all of these contrastive methods require a large set of negative samples, which are sampled from a memory bank that incurs in additional memory and computational costs, which we avoid altogether.

Additional self-supervision tasks have shown strong performance when coupled with representation distillation. Both SSKD~\cite{Xu2020KnowledgeSelf-supervision} and HSAKD~\cite{Yang2021HierarchicalDistillation} introduce auxiliary tasks for classifying image rotation. However, these added self-supervision tasks incur a high training cost due to augmenting the training batches and adding additional classifiers.
Weight sharing through jointly training sub-networks has also been shown to provide implicit knowledge distillation~\cite{Yu2018SlimmableNetworks, Miles2020CascadedSelf-distillation, Yu2019UniversallyTechniques} and promising results.
In this paper, we propose two distinct distillation losses applied to the features before the final fully-connected layer. Similarly to CRD~\cite{Tian2019ContrastiveDistillation}, we posit that the logit representations lack relevant structural information that is necessary for effective distillation through the low dimensional embedding, while using the earlier intermediate representations can hinder the downstream task performance.\par
%
%
\vspace{0.5em}
\noindent\textbf{Information Theory} (IT) provides a natural lens for interpreting and modelling the statistical relationships between intermediate representations of a neural network. This intersection of information theory and deep learning has subsequently led to a rigorous foundation in understanding the dynamics of training~\cite{Advani2019OnLearning, TishbyDeepPrinciple}, while offering fruitful insights into other application domains, such as network pruning and knowledge distillation.
In the context of representation distillation, most losses can be modelled as maximising some lower bound on the mutual information between the student and the teacher~\cite{Tian2019ContrastiveDistillation, Chen2020WassersteinDistillation}. In this work, we propose to forge an alternative connection between knowledge distillation and information theory using infinitely divisible kernels~\cite{Bhati1969InfinitelyMatrices}. Specifically, we show that maximising both the correlation and mutual information yields two complimentary loss functions that can be related to these entropy-like quantities. We achieve this using a matrix-based \functional that closely resembles \Renyi's $\alpha$-entropy~\cite{SanchezGiraldo2013InformationKernels, SanchezGiraldo2015MeasuresKernels, Williams2010NonnegativeInformation}, which is in turn a natural extension of the well-known Shannon's entropy used in IT. More recently, this work has been applied in a representation learning context~\cite{Yu2021DeepFunctional} for parameterising the information bottleneck principle.
%
%
\vspace{-0.5em}
\section{Preliminaries}

 \noindent\textbf{Representation Distillation} describes the methods that use the representation space that is given as the input to the final fully connected layer of a model. The generalised loss used for representation distillation can be concisely expressed in the following form:
%
%
\begin{align}
    \mathcal{L} = &\mathcal{L}_{XE}( \mathbf{y}, \text{softmax}(\mathbf{y}_s) ) + \beta \cdot d(\mathbf{z}_s, \mathbf{z}_t)
\end{align}
%
where $\mathbf{z}_s \in \Real^{d_s}$ and $\mathbf{z}_t \in \Real^{d_t}$ are the student and teacher representations, $\beta$ is a loss weighting, and $d$ is the distillation loss function. 
The cross entropy $\mathcal{L}_{XE}$ between labels $\mathbf{y}$ and student logits $\mathbf{y}_s$ can be defined as the sum of an entropy and KL divergence term. Furthermore, standard KD~\cite{Hinton2013OnLearning} uses a further KL divergence as the distillation loss between the student and teacher logits, with a temperature term to soften or sharpen the two distributions.
 
 Following \cite{Tian2019ContrastiveDistillation}, the motivation for using the feature representation space, as opposed to logits or any of the intermediate feature maps is two-fold. Firstly, this space preserves the structural information about the input, which may be lost in the logits. 
Secondly, intermediate feature matching losses may negatively impact the students' downstream performance in the cross-architecture tasks due to differing inductive biases~\cite{Tian2019ContrastiveDistillation}, while also incurring significant computational and memory overheads due to the high dimensionality of these feature maps.
In our work, to maximize the information transfer, we propose to express the distillation loss $d(.,.)$ as the weighted sum of a correlation and mutual information term. Below we link these two terms to a general formulation of entropy~\cite{SanchezGiraldo2015MeasuresKernels}.\par 
\vspace{-1em}
\paragraph{Information Theory}
\Renyi's $\alpha$-entropy~\cite{Renyi1960OnInformation} 
provides a natural extension of Shannon's entropy, which has been successfully applied in the context of differential privacy~\cite{Mironov2017RenyiPrivacy}, understanding autoencoders~\cite{Yu2019UnderstandingConcepts}, and face recognition~\cite{Shekar2011FaceAnalysis}. For a random variable $X$ with probability density function (PDF) $f(x)$ in a finite set $\rchi$, the $\alpha$-entropy $\textbf{H}_{\alpha}(X)$ is defined as:
\begin{align}
    \textbf{H}_{\alpha}(f) = \frac{1}{1 - \alpha} \log_2 \int_{\chi} f^{\alpha}(x) dx
\end{align}
Where the limit as $\alpha\rightarrow 1$ is the well-known Shannon entropy. 
To avoid the need for evaluating the underlying probability distributions, a set of entropy-like quantities that closely resemble Renyi's entropy were proposed in \cite{SanchezGiraldo2015MeasuresKernels, Williams2010NonnegativeInformation} 
and instead estimate these information quantities directly from data.
They are based on the theory of infinitely divisible matrices and leverage the representational power of reproducing kernel Hilbert spaces (RKHS), which have been widely studied and adopted in classical machine learning. Since its fruition, this framework has been applied in understanding convolutional neural networks (CNNs)~\cite{Yu2020UnderstandingExploration}, whereby they verify the important data processing inequality in information theory and further demonstrate a redundancy-synergy trade-off in layer representations.  We propose to apply these estimators in the context of representation distillation.

We now provide definitions of the entropy-based quantities and their connections with positive semidefinite matrices. This idea then leads to a multi-variate extension using Hadamard products, from which conditional and mutual information can be defined. For brevity, we omit the proofs and connections with \Renyi's axioms, which can be found in \cite{SanchezGiraldo2015MeasuresKernels, Williams2010NonnegativeInformation}.\par
\textit{Definition 1}: Let $X = \{x^{(1)}, \dots x^{(n)}\}$ be a set of $n$ data points of dimension $d$ and $\kappa : X \times X \rightarrow \Real$ be a real-valued positive definite kernel. The Gram matrix $\mathbf{K}$ is obtained from evaluating $\kappa$ on all pairs of examples, that is $K_{ij} = \kappa(x^i, x^j)$. The matrix-based analogue to \Renyi’s $\alpha$-entropy for a normalized positive definite (NPD) matrix $\mathbf{A}$ of size $n \times n$, such that $tr(\mathbf{A}) = 1$, can be given by the following functional: 
\begin{align}
    \label{eqn:entropy}
    \mathbf{S}_{\alpha}(\mathbf{A}) &= \frac{1}{1 - \alpha}\log_{2}(tr(\mathbf{A}^{\alpha})) = \frac{1}{1 - \alpha}\log_{2}\left[\sum_{i=1}^{n}\lambda_{i}(\mathbf{A}^{\alpha})\right]
\end{align}
where $\mathbf{A}$ is the kernel matrix $\mathbf{K}$ normalised to have a trace of $1$ and $\lambda_i(\mathbf{A})$ denotes its $i$-th eigenvalue. 
This estimator can be seen as a statistic on the space computed by the kernel $\kappa$, while also satisfying useful properties attributed to entropy. In practice, the choice of both $\kappa$ and $\alpha$ can be governed by domain-specific knowledge, which we exploit for the task of knowledge distillation. The $log$ in these definitions, conventionally taken as base $2$, can be interpreted as a data-dependant transformation, and its argument is called the \textit{information potential}~\cite{SanchezGiraldo2013InformationKernels}. In an optimisation context, the information potential and entropy definitions can be used interchangeably since they are related by a strictly monotonic function. 


We are interested in the statistical relationship between two sets of variables, namely the student and teacher representations. To measure this relationship, we introduce the notion of joint entropy, which naturally arises using the product kernel.

\textit{Definition 2}: Let $X$ and $Y$ be two sets of data points. After computing the corresponding Gram matrices $\mathbf{A}$ and $\mathbf{B}$, the joint entropy is then given by:
\begin{align}
    \textbf{S}_{\alpha}(\mathbf{A}, \mathbf{B}) = \textbf{S}_{\alpha}\left(\frac{\mathbf{A} \circ \mathbf{B}}{tr(\mathbf{A} \circ \mathbf{B})}\right)
    \label{eqn:joint_entropy}
\end{align}
where $\circ$ denotes the Hadamard product between two matrices. Using these two definitions, the notion of conditional entropy and mutual information can be derived. We focus on the mutual information, which is given by:
\begin{align}
    \textbf{I}_{\alpha}(\mathbf{A} ; \mathbf{B}) = \textbf{S}_{\alpha}(\mathbf{A}) + \textbf{S}_{\alpha}(\mathbf{B}) - \textbf{S}_{\alpha}(\mathbf{A}, \mathbf{B})
    \label{eqn:mutual_info}
\end{align}
Both equation \ref{eqn:joint_entropy} and \ref{eqn:mutual_info} form a foundation for the correlation and mutual information losses respectively, which are proposed in the following section. 
%
%
\section{Information Theoretic Loss Functions}
In this section we introduce two distillation losses that use two distinct and complementary similarity measures between the student and teacher representations. The first loss uses a correlation measure which captures the similarity across the feature dimension, while the second loss is derived from a measure of mutual information and captures the similarity between examples within the mini-batch. 

\subsection{Maximising correlation}

This first loss attempts to correlate the student and teacher representations. The intuition is that if the two sets of representations are perfectly correlated then the student is at least as discriminative as the teacher.
Let $\mathbf{Z}_s \in \Real^{n \times d}$ and $\mathbf{Z}_t \in \Real^{n \times d}$~\footnote{For clarity, we omit a linear embedding layer used on the student representations to match its dimensionality with the teacher.} denote a batch of representations from the student and teacher respectively. These matrices are computed before the final fully-connected layer to preserve the structural information of the data, thus enabling a strong distillation signal for the student.
We first normalise these representations to zero mean and unit variance across the batch dimension and then propose to construct a cross-correlation matrix, $\mathbf{C}_{st} = \mathbf{Z}_s^T\mathbf{Z}_t / n\in \Real^{d \times d}$. Perfect correlation between the two sets of representations is achieved if all of the diagonal entries $v_i = (\mathbf{C}_{st})_{ii}$ are equal to one. 
 To formulate this as a minimization problem, we propose the following loss: \vspace{-0.5em}
\begin{align}
    \label{eqn:correlation_loss}
    \mathcal{L}_{corr} =  \log_2 \sum_{i=1}^d \left|v_i - 1\right|^{2\alpha}
\end{align}

This general objective is motivated by the recent work on Barlow Twins~\cite{Zbontar2021BarlowReduction} for self-supervised learning, however, there are several distinct differences. Firstly, we drop the redundancy reduction term, which minimizes the off-diagonal entries in the cross correlation matrix, since we are not jointly learning both representations, \ie, the teacher is fixed. In fact we observed that this objective significantly hurts the performance of the student. This performance degradation was similarly observed when decorrelating the off-diagonal entries in the self-correlation matrix $\mathbf{C}_{ss}$, and is likely a consequence of the limited model capacity. Secondly, we introduce an $\alpha$ parameter, which provides a natural generalisation to emphasise low or highly correlated features. Finally, the $\log_2$ transformation was empirically shown to improve the performance by reducing spurious variations within a batch. These modifications were not only empirically justified, but also provide a closer relationship with the matrix-based entropy \functional in equation~\ref{eqn:entropy} (see Supplementary).

\vspace{-0.5em}
\subsection{Maximising mutual information}
The correlation loss aims to match the information present in each feature dimension between the teacher and student representations. The mutual information loss provides an additional complimentary objective whereby we transfer the intra-batch similarity (\ie, the relationship between samples) from the teacher representations to the student representations. The natural choice for achieving this through the lens of information theory is to maximise the mutual information between the two representations. Maximising the mutual information has been successfully applied in past distillation methods~\cite{Ahn2019VariationalTransfer}, following the idea that a high mutual information indicates a high dependence between the two models and thus resulting in a strong student representation. Most other works relate their distillation losses to some lower bound on mutual information~\cite{Tian2019ContrastiveDistillation}, however, using an alternative cheap entropy-like estimator, we propose to maximise this quantity directly:
\begin{align}
     \mathcal{L}_{mi} &= -\mathbf{I}_{\alpha}(\mathbf{G}_s ; \mathbf{G}_t) = \textbf{S}_{\alpha}(\mathbf{G}_s, \mathbf{G}_t) - \textbf{S}_{\alpha}(\mathbf{G}_s) - \redcancel{\textbf{S}_{\alpha}(\mathbf{G}_t)}
\end{align}
where $\mathbf{G}_s \in \Real^{n \times n}$ and $\mathbf{G}_t \in \Real^{n \times n}$ are the student and teacher Gram matrices (\ie, $\mathbf{A}$ and $\mathbf{B}$ in equation~\ref{eqn:mutual_info}). These matrices are constructed using a batch of normalised features $\mathbf{Z}_s$ and $\mathbf{Z}_t$ with a polynomial kernel of degree 1. The resulting matrix is subsequently normalised to have a trace of one. The teacher entropy term in this loss is omitted since the teacher weights are fixed during training. Substituting the marginal and joint entropy definitions from equations~\ref{eqn:entropy} and \ref{eqn:joint_entropy}, with $\mathbf{G}_{st} = \mathbf{G}_s \circ \mathbf{G}_t$ (normalised to have a trace of one), leads to
\begin{align}
    \mathcal{L}_{mi} = 
     \frac{1}{1 - \alpha}\log_2\sum_{i=1}^{n} \lambda_i\left(\mathbf{G}_{st}^{\alpha}\right) - \frac{1}{1 - \alpha}\log_2\sum_{i=1}^{n} \lambda_i\left(\mathbf{G}_{s}^{\alpha}\right)
    \label{eqn:mutual_eigen}
\end{align}
Where $\mathbf{G}_{st}$ is also normalised to have unit trace. Since computing the eigenvalues for lots of large matrices can be computationally expensive during training~\cite{Kerr2009QRGPUs}, we restrict our attention to $\alpha = 2$. This allows us to use the Frobenius norm as a proxy objective and one of which has a connection with the eigenspectrum - $\norm{\mathbf{A}_F}^2 = tr(\mathbf{A}\mathbf{A}^H) =  \sum_{i=1}^{n} \lambda_i(\mathbf{A^2})$ since $\mathbf{A}$ is symmetric. 
\begin{align}
     \mathcal{L}_{mi} = \log_2 {\norm{\mathbf{G}_s}_F^2} - \log_2 {\norm{\mathbf{G}_{st}}_F^2}
    \label{eqn:mutual_loss_minimize}
\end{align}
In practice, we observed that removing the $log$ transformations improved the performance, thus resulting in a slight departure from the connection to mutual information. Specifically, the loss instead minimises the distance between the marginal and joint \textit{information potential}, rather than the mutual information (see Supplementary).


\subsection{Combining correlation and mutual information}
Both the proposed losses provide two different learning objectives. Maximising the correlation is applied across the feature dimension, thus ensuring that the students average representation across the batch is perfectly correlated with the teacher. On the other hand, maximising the mutual information encourages the same similarity between samples as from the teacher. These two losses operate distinctly over the two dimensions of the representations, namely the \textit{feature}-dim and the \textit{batch}-dim. The final loss we aim to minimise is given as follows:
\vspace{-0.2em}
\begin{align}
    \mathcal{L}_{ITRD} = \mathcal{L}_{XE} + \beta_{corr}\mathcal{L}_{corr} + \beta_{mi}\mathcal{L}_{mi} 
\end{align}
where $\mathcal{L}_{XE}$ is a cross-entropy loss, while $\beta_{corr}$ and $\beta_{mi}$ are hyperparameters to weight the losses. To demonstrate the simplicity of our proposed method, and similarly to past works \cite{Zbontar2021BarlowReduction}, we provide the PyTorch-based pseudocode in algorithm \ref{alg:training_pseudo_code}.

\renewcommand\thelstlisting{1}

\begin{figure}[!htbp] 
\centering
\begin{minipage}{0.9\linewidth}
\begin{lstlisting}[basicstyle=\scriptsize\ttfamily, mathescape, language=Python, caption={PyTorch-style pseudocode for ITRD}, captionpos=b,
numberbychapter=false,label={alg:training_pseudo_code}]
# f_s: Student network
# f_t: Teacher network
# y: Ground-truth labels
# y_s, y_t: Student and teacher logits
# z_s, z_t: Student and teacher representations (n x d)
for x in loader:
  # Forward pass
  z_s, y_s = f_s(x)
  z_t, y_t = f_t(x)
  z_s = embed(z_s)
  # Cross entropy loss
  loss = cross_entropy(y_s, y)
  
  # Normalise representations
  z_s_norm = (z_s - z_s.mean(0)) / z_s.std(0)
  z_t_norm = (z_t - z_t.mean(0)) / z_t.std(0)
  # Compute cross-correlation vector
  v = einsum('bx,bx$\rightarrow$x', z_s, z_t) / n
  # Compute correlation loss
  dist = torch.pow(v - torch.ones_like(v), 2)
  h_st = torch.log2(torch.pow(dist, alpha).sum())
  loss += h_st.mul(beta_corr)
  
  # Compute Gram matrices
  z_s_norm = normalize(z_s, p=2)
  z_t_norm = normalize(z_t, p=2)
  g_s = einsum('bx,dx$\rightarrow$bd', z_s_norm, z_s_norm)
  g_t = einsum('bx,dx$\rightarrow$bd', z_t_norm, z_t_norm)
  g_st = g_s * g_t
  # Normalize Gram matrices
  g_s = g_s / torch.trace(g_s)
  g_st = g_st / torch.trace(g_st)
  # Compute the mutual information loss
  p = g_s.pow(2) - g_st.pow(2)
  loss += p.sum().mul(beta_mi)
  
  # Optimisation step
  loss.backward()
  optimizer.step()
\end{lstlisting}
\end{minipage}
\end{figure}


%
%
\begin{table*}[t]
    \centering
    \setlength\tabcolsep{5pt}
\resizebox{1.0\textwidth}{!}{\begin{tabular}{c|ccccccc|cccccc}
    \toprule
    \textbf{Teacher} & \cellcolor{sarch}\textbf{W40-2} &  \cellcolor{sarch} \cellcolor{sarch}\textbf{W40-2} &  \cellcolor{sarch}\textbf{R56} &  \cellcolor{sarch}\textbf{R110} &  \cellcolor{sarch}\textbf{R110} &  \cellcolor{sarch}\textbf{R32x4} &  \cellcolor{sarch}\textbf{V13} & \cellcolor{xarch}\textbf{V13} & \cellcolor{xarch}\textbf{R50} & \cellcolor{xarch}\textbf{R50} & \cellcolor{xarch}\textbf{R32x4} & \cellcolor{xarch}\textbf{R32x4} & \cellcolor{xarch}\textbf{W40-2} \\
    \textbf{Student} &  \cellcolor{sarch}\textbf{W16-2} &  \cellcolor{sarch}\textbf{W40-1} &  \cellcolor{sarch}\textbf{R20} &  \cellcolor{sarch}\textbf{R20} &  \cellcolor{sarch}\textbf{R32} &  \cellcolor{sarch}\textbf{R8x4} &  \cellcolor{sarch}\textbf{V8} &  \cellcolor{xarch}\textbf{MN2} & \cellcolor{xarch}\textbf{MN2} & \cellcolor{xarch}\textbf{V8} & \cellcolor{xarch}\textbf{SN1} & \cellcolor{xarch}\textbf{SN2} & \cellcolor{xarch}\textbf{SN1}\\
    \midrule
    Teacher & 75.61 & 75.61 & 72.32 & 74.31 & 74.31 & 79.42 & 74.64 & 74.64 & 79.34 & 79.34 & 79.42 & 79.42 & 75.61 \\ 
    Student & 73.26 & 71.98 & 69.06 & 69.06 & 71.14 & 72.50 & 70.36 & 64.60 & 64.60 & 70.36 & 70.50 & 71.82 & 70.50 \\ 
    \midrule
    KD~\cite{Hinton2015DistillingNetwork} & 74.92 & 73.54 & 70.66 & 70.67 & 73.08 & 73.33 & 72.98 & 67.37 & 67.35 & 73.81 & 74.07 & 74.45 & 74.83\\
    FitNet~\cite{Romero2015FitNets:Nets} & 73.58  & 72.24  & 69.21  & 68.99  & 71.06  & 73.50  & 71.02 & 64.14  & 63.16  & 70.69  & 73.59  & 73.54  & 73.73  \\
    AT~\cite{Zagoruyko2019PayingTransfer} & 74.08  & 72.77  & 70.55  & 70.22  & 72.31  & 73.44  & 71.43 & 59.40  & 58.58  & 71.84  & 71.73  & 72.73  & 73.32  \\
    SP~\cite{Tung2019Similarity-preservingDistillation} & 73.83  & 72.43  & 69.67  & 70.04  & 72.69  & 72.94  & 72.68 & 66.30  & 68.08  & 73.34  & 73.48  & 74.56  & 74.52 \\
    CC~\cite{Peng2019CorrelationDistillation} & 73.56  & 72.21  & 69.63  & 69.48  & 71.48  & 72.97  & 70.71 & 64.86  & 65.43  & 70.25  & 71.14  & 71.29  & 71.38 \\
    RKD~\cite{Park2019RelationalDistillation} & 73.35  & 72.22  & 69.61  & 69.25  & 71.82  & 71.90  & 71.48 & 64.52  & 64.43  & 71.50  & 72.28  & 73.21  & 72.21 \\
    PKT~\cite{Passalis2018LearningTransfer} & 74.54  & 73.45  & 70.34  & 70.25  & 72.61  & 73.64  & 72.88 & 67.13  & 66.52  & 73.01  & 74.10  & 74.69  & 73.89  \\
    FT~\cite{Kim2018ParaphrasingTransfer} & 73.25  & 71.59  & 69.84  & 70.22  & 72.37  & 72.86  & 70.58 & 61.78  & 60.99  & 70.29  & 71.75  & 72.50  & 72.03 \\
    NST~\cite{Huang2017LikeTransfer} & 73.68  & 72.24  & 69.60  & 69.53  & 71.96  & 73.30  & 71.53 & 58.16  & 64.96  & 71.28  & 74.12  & 74.68  & 74.89 \\
    
    CRD~\cite{Tian2019ContrastiveDistillation} & 75.64  & 74.38  & 71.63  & 71.56  & 73.75  & 75.46  & 74.29 & 69.94  & 69.54  & 74.58  & 75.12  & 76.05  & 76.27  \\
    
    
    WCoRD~\cite{Chen2020WassersteinDistillation} & \underline{76.11}  & 74.72  & \textbf{71.92}  & \underline{71.88}  & \underline{74.20}  & \underline{76.15}  & 74.72 & 70.02  & 70.12  & 74.68  & 75.77  & 76.48  & 76.68   \\
    
    ReviewKD~\cite{ChenDistillingReview} & \textbf{76.12}  & \underline{75.09}  & \underline{71.89}  & -  & 73.89  & 75.63  & \underline{74.84} & \underline{70.37}  & 69.89  & - & \textbf{77.45}  & \textbf{77.78}  & \underline{77.14} \\
    
    \midrule
    $\mathcal{L}_{corr}$ & 
    \stackengine{2pt}{75.85}{$_{\pm 0.12}$}{U}{c}{F}{T}{S} & 
    \stackengine{2pt}{74.90}{$_{\pm 0.29}$}{U}{c}{F}{T}{S} & 
    \stackengine{2pt}{71.45}{$_{\pm 0.21}$}{U}{c}{F}{T}{S} & 
    \stackengine{2pt}{71.77}{$_{\pm 0.34}$}{U}{c}{F}{T}{S} & 
    \stackengine{2pt}{74.02}{$_{\pm 0.27}$}{U}{c}{F}{T}{S} &
    \stackengine{2pt}{75.63}{$_{\pm 0.09}$}{U}{c}{F}{T}{S} & 
    \stackengine{2pt}{74.70}{$_{\pm 0.27}$}{U}{c}{F}{T}{S} &  
    \stackengine{2pt}{69.97}{$_{\pm 0.33}$}{U}{c}{F}{T}{S}  & 
    \stackengine{2pt}{\textbf{71.41}}{$_{\pm 0.41}$}{U}{c}{F}{T}{S}  & 
    \stackengine{2pt}{\textbf{75.71}}{$_{\pm 0.02}$}{U}{c}{F}{T}{S} & 
    \stackengine{2pt}{76.80}{$_{\pm 0.28}$}{U}{c}{F}{T}{S} &
    \stackengine{2pt}{77.27}{$_{\pm 0.25}$}{U}{c}{F}{T}{S} &
    \stackengine{2pt}{\textbf{77.35}}{$_{\pm 0.25}$}{U}{c}{F}{T}{S} \\  

    $\mathcal{L}_{corr}+\mathcal{L}_{mi}$ & 
    \stackengine{2pt}{\textbf{76.12}}{$_{\pm 0.04}$}{U}{c}{F}{T}{S} & 
    \stackengine{2pt}{\textbf{75.18}}{$_{\pm 0.22}$}{U}{c}{F}{T}{S} & 
    \stackengine{2pt}{{71.47}}{$_{\pm 0.07}$}{U}{c}{F}{T}{S} & 
    \stackengine{2pt}{\textbf{71.99}}{$_{\pm 0.46}$}{U}{c}{F}{T}{S} & 
    \stackengine{2pt}{\textbf{74.26}}{$_{\pm 0.05}$}{U}{c}{F}{T}{S} &
    \stackengine{2pt}{\textbf{76.19}}{$_{\pm 0.22}$}{U}{c}{F}{T}{S} & 
    \stackengine{2pt}{\textbf{74.93}}{$_{\pm 0.12}$}{U}{c}{F}{T}{S} & 
    \stackengine{2pt}{\textbf{70.39}}{$_{\pm 0.39}$}{U}{c}{F}{T}{S} & 
    \stackengine{2pt}{\underline{71.34}}{$_{\pm 0.33}$}{U}{c}{F}{T}{S}  & \stackengine{2pt}{\underline{75.49}}{$_{\pm 0.32}$}{U}{c}{F}{T}{S}  & 
    \stackengine{2pt}{\underline{76.91}}{$_{\pm 0.19}$}{U}{c}{F}{T}{S} & 
    \stackengine{2pt}{\underline{77.40}}{$_{\pm 0.06}$}{U}{c}{F}{T}{S} &
    \stackengine{2pt}{77.09}{$_{\pm 0.08}$}{U}{c}{F}{T}{S} \\    

    \bottomrule
    \end{tabular}}
    \caption{CIFAR-100 test \textit{accuracy} (\%) of student networks trained with a number of distillation methods. The best results are highlighted in \textbf{bold}, while the second best results are \underline{underlined}. The mean and standard deviation was estimated over 3 runs. Same-architecture transfer experiments are highlighted in blue, whereas cross-architectural transfer is shown in red.}
    \label{table:knowledge_distillation_cifar100_merged}
\end{table*}

\section{Experiments}
We evaluate our proposed distillation across two standard benchmarks, namely the CIFAR-100 and ImageNet datasets. To further demonstrate the effectiveness of our loss, we perform additional experiments on the transferability of the students representations (see Supplementary), distilling from a full-precision model to a binary network, and on an NLP reading comprehension task. 
For all of these experiments, we jointly train the student model with an additional linear embedding for the student representation. This embedding is used for the correlation loss and is shared by the mutual information loss when there is a mismatch in dimensions between the student and the teacher.\par 

\subsection{Model compression}
\textbf{Experiments on CIFAR-100} classification \cite{Krizhevsky2009LearningImages} consist of 60K $32 \times 32$ RGB images across 100 classes with a 5:1 training/testing split. The results are shown in table~\ref{table:knowledge_distillation_cifar100_merged} for multiple student-teacher pairs. For a fair comparison, we include those methods that use the standard CRD~\cite{Tian2019ContrastiveDistillation} teacher weights. The model abbreviations in the results table are given as follows: Wide residual networks (WRNd-w)~\cite{Zagoruyko2016WideNetworks}, MobileNetV2~\cite{Fox2018MobileNetV2:Bottlenecks} (MN2), ShuffleNetV1~\cite{Zhang2018ShuffleNet:Devices} / ShuffleNetV2~\cite{Tan2018MnasNet:Mobile} (SN1 / SN2), and VGG13 / VGG8~\cite{Simonyan2015VeryRecognition} (V13 / V8). R32x4, R8x4, R110, R56 and R20 denote \textbf{CIFAR}-style residual networks, while R50 denotes an \textbf{ImageNet}-style ResNet50~\cite{He2015ResNetRecognition}. CRCD~\cite{ZhuComplementaryDistillation} is not shown in table~\ref{table:knowledge_distillation_cifar100_merged} since it uses different and not publicly available teacher weights\footnote{In addition, using the unofficial code released by the authors, we were unable to replicate their reported results.}.  Although both SSKD and HSAKD do provide official implementations and teacher weights, their use of self-supervision and additional auxiliary tasks is much more computationally expensive and orthogonal to our work. However, we do include these methods in the ImageNet experiment since the same teacher weights are used.\par
\begin{table}
    \centering
    \resizebox{.50\columnwidth}{!}{%
    

\resizebox{1.\textwidth}{!}{\begin{tabular}{c|ccc}
    \toprule
    v.s. & ReviewKD & WCoRD & $\mathcal{L}_{corr}$ \\
    \midrule
    $\mathcal{L}_{corr}$ & \cellcolor{light_red} -3.7\% & \cellcolor{light_green} +16.2\% & -\\
    $\mathcal{L}_{corr}+\mathcal{L}_{mi}$ & \cellcolor{light_green} +6.8\% & \cellcolor{light_green} +24.4\% & \cellcolor{light_green} +10.5\% \\
    \midrule
\end{tabular}}
    }
    \vspace{0.5em}
    \caption{Relative performance improvement (averaged over all architecture pairs in table~\ref{table:knowledge_distillation_cifar100_merged}) of the correlation and mutual information losses against ReviewKD, WCoRD and $\mathcal{L}_{corr}$ only. }
    \label{table:relative_performance}
\end{table}
For all experiments in table~\ref{table:knowledge_distillation_cifar100_merged}, we set $\beta_{corr} = 2.0$ and $\beta_{mi} = 1.0$ (or $\beta_{mi} = 0.0$ when only using $\mathcal{L}_{corr}$). For the correlation loss $\alpha$, we use a value of $1.01$ for the same architectures and $1.50$ for the cross-architectures. ITRD achieves the best performance for 10 out of 13 of the architecture pairs, with a $6.8\%$ and $24.4\%$ relative improvement\footnote{For clarity, we use the same definition for relative improvement as provided in WCoRD~\cite{Chen2020WassersteinDistillation}. This is given by $\frac{X - Y}{X - KD}$, where the $X$ method is compared to $Y$ relative to standard KD with KL divergence.} over ReviewKD and WCoRD respectively. The addition of $\mathcal{L}_{mi}$ is also shown to complement the $\mathcal{L}_{corr}$ loss through a $10.5\%$ average relative improvement over all pairs, as shown in table~\ref{table:relative_performance}.\par
\begin{figure}[t]
\centering
    \includegraphics[width=.50\linewidth]{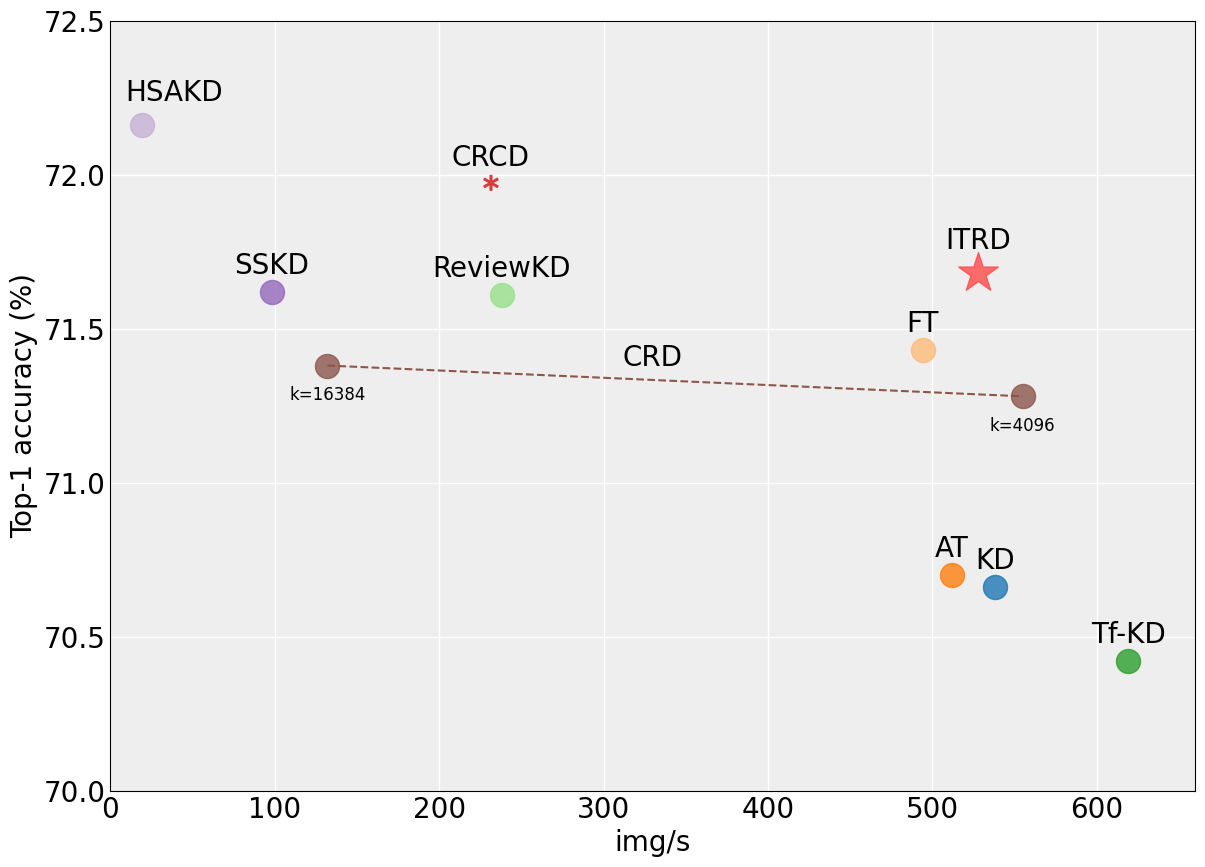}
    \vspace{-0.5em}
    \caption{Top-1 Accuracy on ImageNet \textit{vs} training efficiency with a ResNet-18 as the student and a pre-trained ResNet-34 as the teacher. For CRCD, the training efficiency was evaluated using the authors unofficial implementation, while the accuracy is reported in their paper.}
    \label{fig:training_complexity}
\end{figure}

\noindent\textbf{Experiments on ImageNet} classification~\cite{Russakovsky2014ImageNetChallenge} involve 1.3 million images from  1000 different classes. In this experiment, we set the input size to $224\times224$, and follow a standard augmentation pipeline of cropping, random aspect ratio and horizontal flipping. We use the \textit{torchdistill} library with standard settings, \ie, 100 epochs of training using SGD with an initial learning rate of 0.1 that is divided by 10 at epochs 30, 60 and 90.  The results are shown in figure~\ref{fig:training_complexity} against the total training efficiency, which is measured in \textit{img/s} and is inversely proportional to the total training time. This metric is evaluated using the official \textit{torchdistill} implementations where possible. In the case of HSAKD, we used their official implementation and for CRCD we used the unofficial implementation provided by the authors. For a fair comparison, the batch sizes were scaled to ensure the training would fit within a pre-determined memory constraint of 8GB, and we used for training an RTX 2080Ti GPU. 

In terms of accuracy, ITRD achieves an error of 28.32\%, being only behind CRCD and HSAKD, which are much more computationally costly through the use of either negative contrastive sampling and a gradient-based loss, or additional augmented training data. Conversely, ITRD is computationally efficient, with only a small overhead coming from a single linear layer that embeds the student and teacher representations to the same space, and from computing the gram and cross-correlation matrices. The results show the applicability of ITRD to large-scale datasets, while being significantly more efficient and simple to adopt.

\noindent\textbf{Binary neural networks (BNNs)}  \cite{Ding2019RegularizingNetworks, Qin2020ForwardNetworks, Lin2020RotatedNetwork, Xu2021ReCU:Networks} are an extreme case of quantisation, where the weights can only represent two values. BNNs can obtain a significant model size reduction and increase of inference speed on CPUs~\cite{Rastegari2016XNOR-Net:Networks} and FPGAs~\cite{Umuroglu2017FINN:Inference}, with only a small drop in accuracy. We now show that ITRD can be used to reduce the gap between binary and full-precision (FP) networks. We use the state-of-the-art method ReCU \cite{Xu2021ReCU:Networks} as our base model, and we distill the information from a FP teacher to our BNN student, which share the same architecture apart from the quantisation modules in the student. Table~\ref{table:binary_distillation} shows the results, where for all distillation methods we used the same hyperparameters as in the previous experiments. Both CRD and ReviewKD degrade the BNN performance and, in contrast, ITRD improves upon the original ReCU by $1.3\%$, which is only $0.7\%$ shy of the FP model.\par
\noindent\textbf{NLP Question Answering.} To show the wide applicability of our method, Table~\ref{table:binary_distillation} shows the results of ITRD in a distillation task on the SQuAD 1.1 \cite{rajpurkar2016squad} reading comprehension task, using the transformer-based~\cite{vaswani2017attention} BERT~\cite{devlin-etal-2019-bert} as a teacher and modified versions of BERT with fewer layers as the students. For this experiment, we use the same hyperparameters used in the previous experiments, and following TextBrewer we apply ITRD to the output of each of the student transformer layers, and also use a standard KD~\cite{Hinton2013OnLearning} loss between the teacher and students logits. Table~\ref{table:binary_distillation} shows that we outperform both NLP-specific distillation methods TextBrewer~\cite{Yang2020TextBrewer:Processing} and DistilBert~\cite{DistilBert2019} in both the {Exact Match} (EM) metrics and in F1 score.

\begin{table}[t]
    \centering
    \resizebox{.45\linewidth}{!}{\begin{tabular}{ccc}
    \toprule
    \textbf{Network} & \textbf{Method} &\textbf{ Top-1 (\%)} \\
    \midrule
    ResNet-18 & Full Precision &  94.8 \\
    & RAD~\cite{Ding2019RegularizingNetworks} & 90.5 \\
    & IR-Net~\cite{Qin2020ForwardNetworks} & 91.5 \\
    & RBNN~\cite{Lin2020RotatedNetwork} & 92.2 \\
    & ReCU~\cite{Xu2021ReCU:Networks} &  92.8 \\
    & ReCU + CRD & 92.1 \\
    & ReCU + ReviewKD & 92.6 \\
    & ReCU + KD &  93.3 \\
    & ReCU $+\mathcal{L}_{corr}$ &  93.9 \\
    & ReCU $+\mathcal{L}_{corr}+\mathcal{L}_{mi}$ & \textbf{94.1} \\
    
    
    
    
    
    \bottomrule
\end{tabular}}
\hspace{3em}
\small
\resizebox{.35\linewidth}{!}{\begin{tabular}{cccc}
    \toprule
    & \textbf{Model} & \textbf{EM} & \textbf{F1}  \\
    \midrule
    & Teacher (BERT) & 81.5 & 88.6 \\ 
    \midrule
    \multirow{3}{*}{\begin{sideways} \textbf{T6}\end{sideways}}
    & DistilBERT & 79.1 & 86.9 \\
    & TextBrewer & 80.8 & 88.1 \\
    & ITRD & \textbf{81.5} & \textbf{88.5}\\
    \midrule
    \multirow{2}{*}{\begin{sideways} \textbf{T3}\end{sideways}}
  & TextBrewer & 76.3 & 84.8 \\
  &    ITRD & \textbf{77.7} & \textbf{85.8} \\
    \bottomrule
\end{tabular}}
    \vspace{0.5em}
    \caption{\textbf{Left:} Binary Network classification on CIFAR-10. \textbf{Right:} Question Answering on SQuAD 1.1. The teacher architecture, BERT, contains 12 layers, whereas the students, \textit{T6} and \textit{T3}, follow the same architecture as BERT but with 6 and 3 layers respectively.}
    \label{table:binary_distillation}
\end{table}
\vspace{-0.5em}
%
%
%
%
%
%
%
%
\vspace{-1.em}
\section{Conclusion}
In this work, we proposed an information-theoretic setting for representation distillation. Using this framework, we introduce novel distillation losses that are very simple and computationally inexpensive to adopt into most deep learning pipelines. Each of the proposed losses aims to extract complementary information from the teacher network. The correlation loss guides the student to match the teacher representation on a feature level. Conversely, the mutual information loss transfers the intra-batch similarity between samples from the teacher to the student. We have shown the superiority of our approach compared to methods of similar computational costs on standard classification benchmarks. Furthermore, we have shown the wide applicability of our method by reducing the gap between full-precision and binary networks, and also improving upon NLP-specific distillation methods.\par
\noindent\textbf{Acknowledgement.} This research was supported by UK EPSRC project EP/S032398/1.


\bibliography{references, more_references}

\end{document}